\documentclass[conference]{IEEEtran}
\IEEEoverridecommandlockouts

\usepackage{cite}
\usepackage{amsmath,amssymb,amsfonts}
\usepackage{algorithm, algorithmicx}
\usepackage{graphicx}
\usepackage{textcomp}
\usepackage{xcolor}
\usepackage{booktabs}
\usepackage{subcaption}
\usepackage{algpseudocode}
\usepackage{float}
\usepackage{subcaption}
\def\BibTeX{{\rm B\kern-.05em{\sc i\kern-.025em b}\kern-.08em
    T\kern-.1667em\lower.7ex\hbox{E}\kern-.125emX}}
\begin{document}

\title{Multi-view Clustering via Bi-level Decoupling and Consistency Learning\\
\thanks{$^\star$Corresponding author. This work was funded in part by the National Natural Science Foundation of China (92470202, 62272468 and 62376252); Key Project of Natural Science Foundation of Zhejiang Province (LZ22F030003); Zhejiang Province Leading Geese Plan (2025C02025, 2025C01056); Zhejiang Province Province-Land Synergy Program (2025SDXT004-3).}
}

\author{\IEEEauthorblockN{1\textsuperscript{st} Shihao Dong}
\IEEEauthorblockA{\textit{School of Computer Science} \\
\textit{Nanjing University of Information Science and Technology}\\
Nanjing, China \\
dongshihao@nuist.edu.cn}
\and
\IEEEauthorblockN{2\textsuperscript{nd} Yuhui Zheng}
\IEEEauthorblockA{\textit{The State Key Laboratory of Tibetan Intelligence} \\
\textit{Qinghai Normal University}\\
Xining, China \\
zhengyh@vip.126.com}
\and
\IEEEauthorblockN{3\textsuperscript{rd} Huiying Xu$^\star$}
\IEEEauthorblockA{\textit{Computer Science and Technology} \\
\textit{Zhejiang Normal University}\\
Jinhua, China \\
xhy@zjnu.edu.cn}
\and
\IEEEauthorblockN{4\textsuperscript{th} Xinzhong Zhu}
\IEEEauthorblockA{\textit{Computer Science and Technology} \\
\textit{Zhejiang Normal University}\\
Jinhua, China \\
zxz@zjnu.edu.cn}
}

\maketitle

\begin{abstract}
Multi-view clustering has shown to be an effective method for analyzing underlying patterns in multi-view data. The performance of clustering can be improved by learning the consistency and complementarity between multi-view features, however, cluster-oriented representation learning is often overlooked. In this paper, we propose a novel Bi-level Decoupling and Consistency Learning framework (BDCL) to further explore the effective representation for multi-view data to enhance inter-cluster discriminability and intra-cluster compactness of features in multi-view clustering. Our framework comprises three modules: 1) The multi-view instance learning module aligns the consistent information while preserving the private features between views through reconstruction autoencoder and contrastive learning. 2) The bi-level decoupling of features and clusters enhances the discriminability of feature space and cluster space. 3) The consistency learning module treats the different views of the sample and their neighbors as positive pairs, learns the consistency of their clustering assignments, and further compresses the intra-cluster space. Experimental results on five benchmark datasets demonstrate the superiority of the proposed method compared with the SOTA methods. Our code is published on https://github.com/LouisDong95/BDCL.
\end{abstract}

\begin{IEEEkeywords}
self-supervised learning, deep clustering, multi-view clustering, contrastive learning.
\end{IEEEkeywords}

\section{Introduction}
\label{sec:intro}
In the real world, objects are usually represented by multiple views and perceived by corresponding sensors or receivers. Multi-view clustering (MVC) is widely used to analyze multi-view or multi-modal data to reveal patterns and associations between data. With the development of MVC, some excellent methods have been proposed, including methods based on subspace learning~\cite{gao2015multi, cao2015diversity, kang2020large}, non-negative matrix factorization (NMF)~\cite{liu2013multi, zhao2017multi, huang2020auto} and graph learning~\cite{tang2020cgd, wang2019gmc}. While these traditional methods demonstrate effectiveness on low-dimensional data, they are limited by their data representation capabilities and tend to exhibit diminished performance when confronted with high-dimensional data, such as images and graphs.

\begin{figure}
    \centering
    \includegraphics[width=1\linewidth]{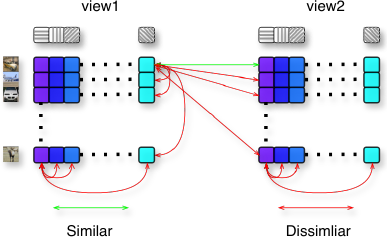}
    \caption{\textbf{Our motivation.} The left and right parts represent the feature matrices of different views, each row is considered as an instance vector and each column is considered as a feature vector.}
    \label{fig_idea}
\end{figure}

In recent years, with the advancement of deep neural networks (DNNs), deep learning methods have been applied to clustering tasks~\cite{xie2016unsupervised, guo2017improved}. Moreover, these techniques have been extended to MVC, which is known as deep multi-view clustering (DMVC)~\cite{abavisani2018deep, li2019deep, xu2021multi, DBLP:conf/icncc/TangTWFW18, cheng2021multi}. In general, DMVC methods follow the principle of complementarity of private features and consistency of consensus features, that is, using an auto-encoder (AE) to learn private features of each view, extracting consistency information through alignment methods such as contrastive learning~\cite{He0WXG20,ChenK0H20}, and further using fused features to guide the learning of each private feature. SDMVC~\cite{xu2022self} uses KL-divergence to align the feature distribution of each view with the distribution of concatenated features. GCFAgg~\cite{yan2023gcfagg} inputs the concatenated features into the attention module to build a global consensus representation, which further guides the view-specific features. CTCC~\cite{dong2023cross} maximizes the mutual information across views to learn consistent information and enhances the complementarity of each view by selectively isolating distributions from each other. DFMVC~\cite{0001PCZCP024} constructs a dynamic graph using latent features, obtains a global representation through the fusion graph, and is used for pseudo-label self-supervised learning to guide the learning of each view. Although these methods effectively separate private features and fuse consistent features, they ignore the inter-cluster distinguishability and intra-cluster compactness of the features themselves, i.e., the degree of distinction between features of different clusters and the distribution of features of the same cluster in space, both of them play a key role in clustering performance.

In this paper, we propose a novel framework of bi-level decoupling and consistency learning multi-view clustering (BDCL) to address the above problem. As shown in Fig.~\ref{fig_idea}, contrastive learning regards different views of the same sample as positive pairs and the remaining views as negative pairs to learn discriminative representations of instances. Inspired by contrastive learning, we treat the feature of each dimension in each view as a separate class, and each feature should be similar to itself and distinct from the others. In image data, these features usually contain information about objects, such as color, edges, etc., but coupled features will cause some information to overlap, resulting in a reduction of the information contained in the features. To retain the feature information of each instance to the greatest extent, we decouple the embedding features at the feature level. Meanwhile, the coupling of cluster features will cause each class to be indistinguishable in space. Therefore we further decouple the cluster features at the cluster level to improve the discriminability of clusters. In addition, to improve the compactness of the cluster space, we consider different views of instance and their neighbors as positive pairs and learn the consistent clustering assignment between them through the consistency learning module. In summary, we first maximize the preservation of private information of the data through the reconstruction of AE, and then align the consensus embedding features through contrastive learning. The clustering consistency learning module implements clustering assignment and compresses the cluster space. At the same time, the embedding features and cluster features of each view are decoupled, respectively, to further improve the discriminative ability of features and clusters. The final results of the clustering are obtained through the clustering assignment. The main contributions are summarized as follows:
\begin{itemize}
\item We propose a novel end-to-end DMVC framework, termed BDCL, which combines multi-view consistency learning with feature decoupling.
\item We propose a novel bi-level decoupling method, where feature-level decoupling and cluster-level decoupling effectively improve the discriminability of embedded features and the discriminability of cluster features.
\item Extensive experiments conducted on five multi-view datasets demonstrate the effectiveness of our method. 
\end{itemize}

\section{Related works}
\subsection{Multi-view Clustering}
MVC not only considers data from different perspectives and data from different modalities, but also fuses features from different perspectives to extract consistent and complementary features from different perspectives. According to the feature extraction method, MVC includes traditional MVC and DMVC. Although many excellent algorithms have been proposed, traditional methods face the problem of dimensionality disaster.

In recent years, DMVC has been greatly developed due to the powerful feature extraction ability of DNNs. According to the clustering method can be divided into three categories: 1) DEC-based: DAMC~\cite{li2019deep} proposed a deep adversarial MVC network to learn the intrinsic structure embedding in multi-view data. AIMC~\cite{xu2019adversarial} extended it to handle incomplete multi-view data. Multi-VAE~\cite{xu2021multi} proposed a VAE-based MVC framework by learning disentangled visual representations. 2) Subspace clustering-based: DMSC~\cite{abavisani2018deep} inserted self-expression layers in AEs and compared the clustering performance of the fusion methods at different stages. DMVSSC~\cite{DBLP:conf/icncc/TangTWFW18} proposed a deep multi-view sparse subspace clustering model consisting of convolutional-AE and self-expressive module. 3) GNN-based: MAGCN~\cite{cheng2021multi} proposed a multi-view attribute graph convolution networks model for handling node attributes and graph reconstruction. DFMVC~\cite{0001PCZCP024} proposed a dynamic graph fusion method to guide the learning of each view feature through the fused consistent features.

\subsection{Feature Decoupling}

The feature coupling problem refers to the problem that there is a strong correlation or dependency between different features, which has a negative impact on the performance, interpretability or generalization ability of the model. Feature coupling includes instance-level coupling and cluster-level coupling. Existing works have sought to address this issue by decoupling features to enhance their representational power. For example, IDFD~\cite{TaoTN21} reduced the correlation of instances and features through contrastive learning. AGC-DRR~\cite{GongZTL22} perturbed the original graph to reduce the redundancy in the input space, and the association matrix and the unit matrix are close to reduce the redundancy in the feature space. DCRN~\cite{LiuTZLSYZ22} avoided clustering collapse by reducing the correlation of instance and cluster features respectively. Although the redundancy between instances can reduce the dependence between negative samples with the help of contrastive learning~\cite{ChenK0H20,He0WXG20}, few works focus on the coupling of embedded features and cluster features at the same time. 

In contrast to these methods, our bi-level decoupling approach simultaneously considers the coupling at both the feature level and the cluster level, offering a more comprehensive solution to the feature coupling problem.

\begin{figure*}
    \centering
    \includegraphics[width=1\linewidth]{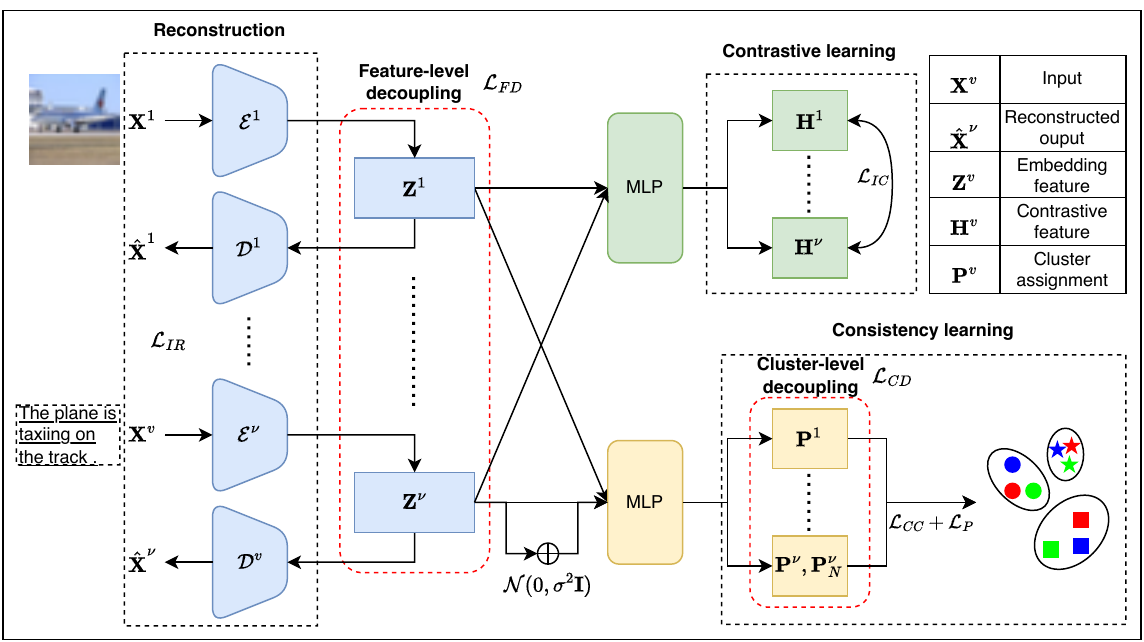}
    \caption{\textbf{The framework of BDCL.} AE learns embedding features $\mathbf{Z}^\nu$ through reconstructed data $\hat{\mathbf{X}}^\nu$ and original data $\mathbf{X}^\nu$. Input the  $\mathbf{Z}^\nu$ into the contrastive MLP to obtain the contrastive feature $\mathbf{H}^\nu$, and perform contrastive learning, which can increase the discriminability of the instance. Input the $\mathbf{Z}^\nu$ and neighbor features of different views into the clustering MLP to learn clustering consistency and effectively compress the cluster space. Feature-level and cluster-level decoupling further improves the discriminability of embedding features and cluster features.}
    \label{fig_pipline}
\end{figure*}

\section{Method}
In this section, we proposed a novel framework of BDCL to explore the feature and cluster bi-level decoupling and clustering consistency learning how to improve multi-view representation and enhance clustering performance. Fig.~\ref{fig_pipline}. illustrates our framework.

\subsection{Notation and Problem}

\begin{table}
    \caption{Notation Summary}
    \centering
    \begin{tabular}{ll}
    \toprule
        Notation & Meaning\\
    \midrule
        $\mathbf{X}^{\nu}\in\mathbb{R}^{N\times d_{\nu}}$ & The $\nu$-th view of dataset\\
        $\mathbf{Z}^{\nu}\in\mathbb{R}^{N\times m}$ & The $\nu$-th view of embedding features\\
        $\mathbf{H}^{\nu}\in\mathbb{R}^{N\times q}$ & The $\nu$-th view of contrastive features\\
        $\mathbf{P}^{\nu}\in\mathbb{R}^{N\times K}$ & The $\nu$-th view of clustering assignment\\
        $\mathbf{x}_i^{\nu}\in\mathbb{R}^{d_{\nu}}$ & The $i$-th sample of $\nu$-th view\\
        $\mathbf{z}_i^{\nu}\in\mathbb{R}^m$ & The $i$-th embedding instance of $\nu$-th view\\
        $\mathbf{h}_i^{\nu}\in\mathbb{R}^q$ & The $i$-th contrastive instance of $\nu$-th view\\
        $\mathbf{p}_i^{\nu}\in\mathbb{R}^K$ & The $i$-th clustering assignment of $\nu$-th view\\
        $\mathcal{E}^{\nu}, \mathcal{D}^{\nu}$ & Encoder and decoder of $\nu$-th view\\
        $\tau, \sigma, \lambda$ & hyper-parameter\\
    \bottomrule
    \end{tabular}
    \label{tab:my_label}
\end{table}

Given a set of multi-view dataset $\mathcal{X}=\{\mathbf{X}^{\nu}\in \mathbb{R}^{N\times d_{\nu}}\}_{{\nu}=1}^V$ with $V$ views and $N$ samples. The dimension of each sample $\mathbf{x}_i^{\nu}$ in ${\nu}$-th view $\mathbf{X}^{\nu}=[\mathbf{x}^{\nu}_1, \mathbf{x}^{\nu}_2, \dots, \mathbf{x}^{\nu}_N]^\top$ is $d_{\nu}$. The task of MVC is to divide the dataset $\mathcal{X}$ into specified $K$ clusters without labels.

\subsection{Multi-view Instance Learning}
The raw data $\mathcal{X}$ is usually high-dimensional and indistinguishable, the general method reconstructs the original data $\mathbf{X}^{\nu}$ of $\nu$-th view through encoder $\mathcal{E}^{\nu}$ and decoder $\mathcal{D}^{\nu}$ of $\nu$-th AE, and learns the embedding features $\mathbf{Z}^{\nu} \in \mathbb{R}^{N\times m}$ of each view, $\mathbf{Z}^{\nu} = \mathcal{E}^{\nu}(\mathbf{X}^{\nu})$, which maximize the preservation of the information of the original data. The reconstruction loss is defined as follows:
\begin{equation}\label{eq_reconstruction}
    \mathcal{L}_{IR} = \frac{1}{N}\sum_{\nu=1}^V\sum_{i=1}^N\parallel \mathbf{x}_i^{\nu}-\hat{\mathbf{x}}_i^{\nu}\parallel_2^2,
\end{equation}
where reconstructed data of $\nu$-th view is $\hat{\mathbf{X}}^{\nu} = \mathcal{D}^{\nu}(\mathbf{Z}^{\nu})$. Although the dimensionality of the original data is effectively reduced by the encoder and retaining information, the embedding features are not clustering oriented, and performing clustering directly on these features does not achieve good performance. Some recent work~\cite{xu2022multi, yan2023gcfagg} improved instance discrimination by combining AE and contrastive learning. So we map the embedding features into the contrastive space $\mathbf{H}^{\nu} \in \mathbb{R}^{N\times q}$ via MLP and perform contrastive learning of instance in that space. The instance contrastive loss is defined as follows:
\begin{equation}\label{eq_contrastive}
    \mathcal{L}_{IC} = -\frac{1}{2N}\sum_{\substack{u, \nu\\ u\neq \nu}}^V\sum_{i=1}^N\log \frac{e^{s(\mathbf{h}^u_i, \mathbf{h}^{\nu}_i)/\tau}}{\sum_{j=1, j\neq i}^N e^{s(\mathbf{h}^u_i, \mathbf{h}^u_j)/\tau}+e^{s(\mathbf{h}^u_i, \mathbf{h}^{\nu}_j)/\tau}},
\end{equation}
where $\tau$ is the temperature parameter used to scale the similarities between instances, $s(\cdot, \cdot)$ is a similarity measure function used to compute the similarity between two instances, usually calculated by cosine similarity:
\begin{equation}
    s(\mathbf{h}^u_i, \mathbf{h}^{\nu}_i) = \frac{\mathbf{h}^u_i\cdot \mathbf{h}^{\nu}_i}{\parallel \mathbf{h}^u_i\parallel\parallel \mathbf{h}^{\nu}_i\parallel}.
\end{equation}
Compared with the contrastive features $\mathbf{H}^{\nu}$, although the embedding features contain much more information $(q < m)$, performing contrastive learning directly on the embedded feature will lead to the conflict between reconstruction and discriminative learning. Therefore, mapping the embedding features to the contrastive space through the MLP for instance contrastive learning not only avoids the loss of information, but also improves the discriminative ability of the instances in the embedding features.

\subsection{Clustering Consistency Learning}
Since embedding features contain more information, we perform clustering on them for better performance. The usual methods fuse the embedding features of each view to learn consistent features and perform clustering on those features to get the final result, however, the quality of the features is different for each view and simple fusion will lead to degradation of the quality of the fused features. We map the embedding features into the cluster space $\mathbf{P}^{\nu} \in \mathbb{R}^{N\times K}$ to avoid the negative effects of feature fusion through clustering consistency learning. Specifically, the same samples $\mathbf{z}_i^u$ and $\mathbf{z}_i^{\nu}$ from different views describe the same object, which should be semantically the same. We use the clustering MLP to map it into the cluster space $\mathbf{p}_i^u$ and $\mathbf{p}_i^{\nu}$, so their clustering assignments should be similar. To make the cluster space distribution more compact, we further learn consistent assignments between instances and their neighboring samples. Neighbor features are represented by instance features with added noise:
\begin{equation}\label{eq_neighbor}
    N(\mathbf{Z}^\nu) = \mathbf{Z}^\nu + \sigma\boldsymbol{E}, \quad \rm{with} \quad\boldsymbol{E}\sim\mathcal{N}(0, \mathbf{I}),
\end{equation}
where $N(\mathbf{Z}^\nu)$ denotes the neighbor features of $\nu$-th embedding features, $\mathbf{I}$ denotes the identity matrix and $\sigma$ is a hyperparameter used to control the range of neighbors. We consider the same sample in different views and neighbor samples as positive pairs and keep them close in cluster space. The clustering consistency loss is defined as follows:
\begin{equation}
    \mathcal{L}_{CC} = \frac{1}{4N}\sum_{\substack{u, \nu\\ u\neq \nu}}^V\sum_{i=1}^N(\parallel \mathbf{p}_i^u - \mathbf{p}_i^{\nu}\parallel_2^2 + \parallel \mathbf{p}_i^u - \mathbf{p}_{N_i}^{\nu}\parallel_2^2),
\end{equation}
where $\mathbf{p}_{N_i}^\nu$ denotes the clustering assignment of the neighbor of $i$-th sample of the $\nu$-th view.

To avoid the collapse of clustering, where all samples are classified in the same cluster, we add a regularization term constraining the cluster space so that the samples are uniformly distributed in the cluster space. The regularization term is defined as follows:
\begin{equation}
    \mathcal{L}_P = \sum_{\nu =1}^V\sum_{j=1}^K{\mathbf{p}'}^{\nu}_j\log {\mathbf{p}'}_j^{\nu},\quad
    {\rm with} {\quad\mathbf{p}'}^{\nu}_j = \frac{1}{N}\sum_{i=1}^N \mathbf{p}^{\nu}_{ij}.
\end{equation}

\subsection{Bi-level Decoupling}
Clustering performance can be enhanced through instance learning and consistency learning. However, these methods primarily focus on the discriminability at the instances, overlooking the discriminability of features. The coupling of embedding features leads to a reduction in the amount of useful information contained in the features. The coupling of cluster features leads to unclear delineation of each cluster to the extent that multiple classes overlap in the spatial distribution. Therefore, we propose a bi-level decoupling to reduce the coupling of features and clusters. Feature-level decoupling makes $m$-dimensional embedding features as independent as possible from each other and distinguished from features in other dimensions. Cluster-level decoupling makes $K$-dimensional clustering features describe unique clusters as much as possible and distinguish them from other clusters. The bi-level decoupling loss consists of the feature-level decoupling loss $\mathcal{L}_{FD}$ and the cluster-level decoupling loss $\mathcal{L}_{CD}$, which is defined as follows:
\begin{equation}\label{eq_loss_dd}
\begin{aligned}
    \mathcal{L}_{BD} &= \mathcal{L}_{FD} + \mathcal{L}_{CD}\\
    &= \sum_{\nu=1}^V \frac{1}{m^2}\parallel {\overline{\mathbf{Z}}^{\nu}}^\top\overline{\mathbf{Z}}^{\nu} - \mathbf{I}_m\parallel_2^2 + \frac{1}{K^2}\parallel {\overline{\mathbf{P}}^{\nu}}^\top\overline{\mathbf{P}}^{\nu} - \mathbf{I}_K\parallel_2^2.
\end{aligned}
\end{equation}
We consider the column vectors of the embedded features $\mathbf{Z}^{\nu}$ of each view as features of each dimension and each column of the assignment matrix $\mathbf{P}^{\nu}$ as features of each class. The similarity between each feature is obtained by the inner product of the corresponding two features after normalization ${\overline{\mathbf{Z}}^{\nu}}^\top\triangleq{\mathbf{Z}^{\nu}}^\top/\parallel{\mathbf{Z}^{\nu}}^\top\parallel_2, {\overline{\mathbf{P}}^{\nu}}^\top\triangleq{\mathbf{P}^{\nu}}^\top/\parallel{\mathbf{P}^{\nu}}^\top\parallel_2$. The similarity between the same features is $1$, and the similarity between different features is close to $0$, making different features independent of each other, thereby improving the representation and distinction capabilities between features.

\subsection{Loss Function}
In summary, we present the entire BDCL framework. In the training phase we need to jointly optimize the instance learning loss $\mathcal{L}_I$, the clustering loss $\mathcal{L}_C$ and the bi-level decoupling loss $\mathcal{L}_{BD}$. The final loss function is defined as follows:
\begin{equation}\label{eq_total_loss}
\begin{aligned}
    \mathcal{L} = &\mathcal{L}_I + \lambda_1\mathcal{L}_C + \lambda_2\mathcal{L}_{BD},\\
    with\quad&\mathcal{L}_I = \mathcal{L}_{IR} + \mathcal{L}_{IC}, \\
    &\mathcal{L}_C = \mathcal{L}_{CC} + \mathcal{L}_P.
\end{aligned}  
\end{equation}
where $\lambda_1, \lambda_2$ are trade-off hyper-parameters. The whole procedure is summarized in Algorithm~\ref{alg}.

\subsection{Complexity Analysis}
In this section, we consider the complexity in Algorithm~\ref{alg}. Here, $B$ denotes the Batch size.
$\mathcal{O}(VBd_{\nu})$, $\mathcal{O}(V^2qB^2)$, $\mathcal{O}(V^2BK)$, $\mathcal{O}(VBm^2)$, $\mathcal{O}(VBK^2)$ are the complexity of reconstruction, instance contrastive learning, cluster consistency learning, feature-level decoupling and cluster-level decoupling respectively. Since $V, K < B < m$, the total time complexity of BDCL is $\mathcal{O}(VBm^2)$. Besides, the space complexity of BDCL is $\mathcal{O}(m^2)$.

\begin{algorithm}\label{alg}
    \caption{BDCL}
    \label{alg}
    \begin{algorithmic}[1]
        \Statex \textbf{Input:} Multi-view dataset $\{\mathbf{X}^\nu\}_{\nu=1}^V$; Number of clustes $K$; Temperature parameters $\tau$; Hyper-parameter $\sigma$; Pre-training epoch $T_1$ and clustering epoch $T_2$.
        \Statex \textbf{Output:} Clustering result.
        \For{epoch = 1 to $T_1$}
            \State Pre-training $\{\mathcal{E}^\nu, \mathcal{D}^\nu\}_{\nu=1}^V$ by minimizing Eq. \eqref{eq_reconstruction};
        \EndFor
        \For{epoch = 1 to $T_2$}
            \State Computing reconstructed data $\{\hat{\mathbf{X}}^{\nu}\}_{\nu=1}^V$;
            \State Computing embedding features $\{\mathbf{Z}_\nu^V\}_{\nu=1}^V$;
            \State Computing neighbor features by eq.~\eqref{eq_neighbor};
            \State Computing contrastive features $\{\mathbf{H}_\nu^V\}_{\nu=1}^V$;
            \State Computing clustering assignment $\{\mathbf{P}_\nu^V\}_{\nu=1}^V$;
            \State Optimizing the total network by minimizing Eq. \eqref{eq_total_loss};
        \EndFor
        \State Computing the final clustering assignment.
    \end{algorithmic}
\end{algorithm}

\begin{table}
    \caption{Summary of datasets}
    \centering
    \setlength{\tabcolsep}{5pt}
    \begin{tabular}{lllcc}
    \toprule
        Datasets & Samples & Dimensions & Views & Classes\\
    \midrule
        MNIST-USPS & 5,000 & 784/784 & 2 & 10\\
        BDGP & 2,500 & 1,750/79 & 2 & 5\\
        CCV & 6,773 & 5,000/5,000/4,000 & 3 & 20\\
        Fashion & 10,000 & 784/784/784 & 3 & 10\\
        Caltech-2V & 1,400 & 40/254 & 2 & 7\\
        Caltech-3V & 1,400 & 40/254/928 & 3 & 7\\
        Caltech-4V & 1,400 & 40/254/928/512 & 4 & 7\\
        Caltech-5V & 1,400 & 40/254/928/512/1,984 & 5 & 7\\
    \bottomrule
    \end{tabular}
    \label{tab_datasets}
\end{table}

\section{Experiment}

\begin{table*}
    \caption{Clustering performance comparison on four datasets.}
    \centering
    \begin{tabular}{lcccccccccccc}
    \toprule
        Datasets & \multicolumn{3}{c}{MNIST-USPS} & \multicolumn{3}{c}{BDGP} & \multicolumn{3}{c}{CCV} & \multicolumn{3}{c}{Fashion}\\
        \cmidrule(lr){2-4} \cmidrule(lr){5-7} \cmidrule(lr){8-10} \cmidrule(lr){11-13}
         Evaluation metrics & ACC & NMI & PUR & ACC & NMI & PUR & ACC & NMI & PUR & ACC & NMI & PUR\\
    \midrule
        CDIMC-net~\cite{wen2021cdimc} & 0.620 & 0.676 & 0.647 & 0.884 & 0.799 & 0.885 & 0.201 & 0.171 & 0.218 & 0.776 & 0.809 & 0.789\\
        EAMC~\cite{zhou2020end} & 0.735 & 0.837 & 0.778 & 0.681 & 0.480 & 0.697 & 0.263 & 0.267 & 0.274 & 0.614 & 0.608 & 0.638\\
        SiMVC~\cite{trosten2021reconsidering} & 0.981 & 0.962 & 0.981 & 0.704 & 0.545 & 0.723 & 0.151 & 0.125 & 0.216 & 0.825 & 0.839 & 0.825\\
        CoMVC~\cite{trosten2021reconsidering} & 0.987 & 0.976 & 0.989 & 0.802 & 0.670 & 0.803 & 0.296 & 0.286 & 0.297 & 0.857 & 0.864 & 0.863\\
        DSMVC~\cite{tang2022deep} & 0.963 & 0.943 & 0.963 & 0.758 & 0.614 & 0.758 & - & - & - & 0.896 & 0.868 & 0.896\\
        MFL~\cite{xu2022multi} & {0.996} & {0.988} & {0.996} & \underline{0.992} & 0.972 & \underline{0.992} & 0.297 & {0.302} & 0.338 & \textbf{0.994} & \textbf{0.984} & \textbf{0.994}\\
        CVCL~\cite{Chen_2023_ICCV} & \underline{0.997} & {0.991} & \underline{0.997} & \underline{0.992} & {0.973} & \underline{0.992} & {0.310} & 0.284 & {0.343} & \underline{0.993} & 0.982 & \underline{0.993}\\
        CSOT~\cite{ZhangZSCLZ24} & \textbf{0.998} & \textbf{0.995} & \textbf{0.998} & \underline{0.992} & \underline{0.975} & \underline{0.992} & \underline{0.327} & \underline{0.310} & \underline{0.356} & \underline{0.993} & \underline{0.983} & \underline{0.993}\\
        BDCL (ours) & \textbf{0.998} & \underline{0.993} & \textbf{0.998} & \textbf{0.994} & \textbf{0.980} & \textbf{0.994} & \textbf{0.330} & \textbf{0.323} & \textbf{0.379} & \textbf{0.994} & \textbf{0.984} & \textbf{0.994}\\
    \bottomrule
    \end{tabular}
    \label{tab_result1}
\end{table*}

\begin{table*}
    \caption{Clustering performance comparison on Caltech with different views.}
    \centering
    \begin{tabular}{lcccccccccccc}
    \toprule
        Datasets & \multicolumn{3}{c}{Caltech-2V} & \multicolumn{3}{c}{Caltech-3V} & \multicolumn{3}{c}{Caltech-4V} & \multicolumn{3}{c}{Caltech-5V}\\
        \cmidrule(lr){2-4} \cmidrule(lr){5-7} \cmidrule(lr){8-10} \cmidrule(lr){11-13}
         Evaluation metrics & ACC & NMI & PUR & ACC & NMI & PUR & ACC & NMI & PUR & ACC & NMI & PUR\\
    \midrule
        CDIMC-net~\cite{wen2021cdimc} & 0.515 &0.480 &0.564 &0.528 &0.483 &0.565 &0.560 &0.564 &0.617 &0.727 &0.692 &0.742\\
        EAMC~\cite{zhou2020end} & 0.419 &0.256 &0.427 &0.389 &0.214 &0.398 &0.356 &0.205 &0.370 &0.318 &0.173 &0.342\\
        SiMVC~\cite{trosten2021reconsidering} & 0.508& 0.471 &0.557 &0.569 &0.495 &0.591 &0.619 &0.536& 0.630 &0.719 &0.677& 0.729\\
        CoMVC~\cite{trosten2021reconsidering} & 0.466 &0.426& 0.527 &0.541& 0.504 &0.584 &0.568 &0.569 &0.646& 0.700 &0.687 &0.746\\
        DSMVC~\cite{tang2022deep} & 0.533 & 0.392 & 0.533 & {0.622} & {0.555} & {0.660} & 0.767 & \underline{0.724} & 0.784 & {0.841} & {0.741} & {0.841}\\
        MFL~\cite{xu2022multi} & 0.606 & 0.528 & 0.616 & 0.631 & 0.566 & 0.639 & 0.733 & 0.652 & 0.734 & 0.804 & 0.703 & 0.804\\
        CVCL~\cite{Chen_2023_ICCV} & \underline{0.650} & {0.537} & \underline{0.654} & \underline{0.706} & {0.613} & \underline{0.716} & {0.790} & 0.664 & {0.790} & 0.820 & 0.703 & 0.820\\
        CSOT~\cite{ZhangZSCLZ24} & 0.639 & \textbf{0.554} & 0.642 & 0.705 & \underline{0.623} & 0.715 & \underline{0.812} & 0.690 & \underline{0.812} & \underline{0.882} & \underline{0.793} & \underline{0.882}\\ 
        BDCL (ours) & \textbf{0.670} & \underline{0.544} & \textbf{0.670} & \textbf{0.746} & \textbf{0.654}& \textbf{0.748} & \textbf{0.839} & \textbf{0.758} & \textbf{0.839} & \textbf{0.898} & \textbf{0.831} & \textbf{0.898}\\
    \bottomrule
    \end{tabular}
    \label{tab_result2}
\end{table*}
\subsection{Datasets and Evaluation Metrics}
In the experiments, we adopt five widely used multi-view datasets to verify the performance of our method.  MNIST-USPS~\cite{peng2019comic} dataset contains 5,000 samples with two different styles of digital images.
BDGP~\cite{cai2012joint} dataset contains 2,500 samples of drosophila embryos, each of which is represented by visual and textual features.
CCV~\cite{jiang2011consumer} dataset is a video dataset with 6,773 samples belonging to 20 classes and provides hand-crafted Bag-of-Words representations of three views, such as STIP, SIFT, and MFCC.
Fashion~\cite{xiao2017fashion} dataset contains 10,000 images of products. Each image is represented by three different styles.
Caltech~\cite{fei2004learning} is an object recognition dataset containing 1,400 images with 7 distinct classes, each image is characterized by three features including WM, CENTRIST, LBP, GIST and HOG. The dataset is summarized in Table~\ref{tab_datasets}.

There are three metrics are widely used in MVC, including clustering accuracy (ACC), normalized mutual information (NMI) and purity (PUR). Higher values of these indicate better clustering performance. Reported values in all experiments are averaged over multiple runs of results, more than three times, to ensure accurate results.

\subsection{Implementation Details}
In our paper, all experiments are conducted on the desktop computer with the Intel Core i7-12700KF CPU, one NVIDIA RTX 3080TI GPU, 32GB RAM, Python3.7 and Pytorch deep learning platform. For the SOTA method we used their configuration and source code and reproduced their results as a comparison, for the other methods we used the results reported on~\cite{xu2022multi}. The proposed network architecture consists of reconstruction module, contrastive module and clustering module. The reconstruction module consists of $V$ AEs with different structures and weights, the dimension of the embedding features are all 512. the contrastive MLP consists of a single one-layer FC layer with output dimension 128. the clustering MLP consists of a single one-layer FC layer with output dimension $K$. $\tau = 0.5, \lambda_1=1, \lambda_2=1, \sigma=0.001$. For a fair comparison, the results are obtained by executing the program multiple times and averaging them.

\begin{table}
    \caption{The impact of different modules on the results.}
    \centering
    \setlength{\tabcolsep}{2pt}
    \begin{tabular}{lcccccccc}
    \toprule
        Datasets & \multicolumn{2}{c}{MNIST-USPS} & \multicolumn{2}{c}{BDGP} & \multicolumn{2}{c}{CCV} & \multicolumn{2}{c}{Fashion}\\
        \cmidrule(lr){2-3}\cmidrule(lr){4-5}\cmidrule(lr){6-7}\cmidrule(lr){8-9}
        Loss & ACC & NMI & ACC & NMI  & ACC & NMI & ACC & NMI\\
    \midrule
        Baseline + $k$-means & 0.983 & 0.963 & 0.982 & 0.947 & 0.311 & 0.308 & {0.991} & {0.977}\\

        BDCL W/O $\mathcal{L}_{BD}$ & 0.993 & 0.979 & {0.992} & 0.974 & 0.303 & 0.307 & \underline{0.993} & \underline{0.981}\\
        
        BDCL W/O $\mathcal{L}_{FD}$ & {0.995} & {0.986} & \underline{0.993} & \underline{0.977} & 0.292 & 0.302 & 0.992 & 0.980\\
        
        BDCL W/O $\mathcal{L}_{CD}$ & \underline{0.997} & \underline{0.991} & {0.992} & 0.975 & \underline{0.312} & \underline{0.315} & \underline{0.993} & \underline{0.981}\\
        
        BDCL & \textbf{0.998} & \textbf{0.993}  & \textbf{0.994} & \textbf{0.980} & \textbf{0.330} & \textbf{0.323} & \textbf{0.994} & \textbf{0.984}\\
    \bottomrule
    \end{tabular}
    \label{tab_ablation1}
\end{table}

\subsection{Performance Comparison}
We compared multiple methods on five datasets, and the results are presented in Table.~\ref{tab_result1} and Table.~\ref{tab_result2}. These methods includes CDIMC-net~\cite{wen2021cdimc}, EAMC~\cite{zhou2020end}, SiMVC~\cite{trosten2021reconsidering}, CoMVC~\cite{trosten2021reconsidering}, DSMVC~\cite{tang2022deep}, MFL~\cite{xu2022multi}, CVCL~\cite{Chen_2023_ICCV} and CSOT~\cite{ZhangZSCLZ24}. Notably, BDCL achieves SOTA results on all five datasets. This is primarily attributed to the enhanced discriminability of both the feature space and the cluster space, facilitated by the bi-level decoupling, along with the compact distribution of cluster space achieved through consistency learning. Despite MFL, CVCL and CSOT utilizing contrastive learning within the cluster space to enhance the discriminability of each cluster, our method consistently outperforms them.

In addition, we conducted comparative experiments on Caltech with a different number of views. As shown in Table.~\ref{tab_result2}, the performance of some methods decreases with the increase in the number of views, because forcing alignment of private view features may lead to the inclusion of information that is not conducive to clustering, thereby harming the fused information. While BDCL specifically learns the consistency of cluster assignments during the clustering stage, making it unaffected by changes in the number of views. This inherent characteristic gives our method a degree of robustness, allowing it to achieve optimal performance across a varying number of views.

\subsection{Ablation Analysis}
To further understand the impact of the combination of loss functions on the final results, we conducted an ablation experiment on the Eq.~\eqref{eq_total_loss}, and the results are shown in Table.~\ref{tab_ablation1}. We use the combination of reconstruction and contrastive learning as the baseline, i.e., the multi-view instance learning module, and its results are obtained by performing $k$-means on each view and averaging the results. The cluster consistency learning module has limited effect in improving clustering performance, mainly because the simple consistency learning module cannot effectively map the embedded features into the cluster space, resulting in a lack of distinguishability within the cluster space. Moreover, compared with cluster-level decoupling, since the dimension of the feature space is much larger than that of the cluster space, the clustering performance improvement brought by feature-level decoupling is more significant on most datasets. Overall, the combination of cluster consistency learning module and bi-level decoupling shows the best performance on all datasets.

\begin{figure}
    \centering
    \begin{subfigure}[b]{0.49\linewidth}
        \centering
        \includegraphics[width=\linewidth]{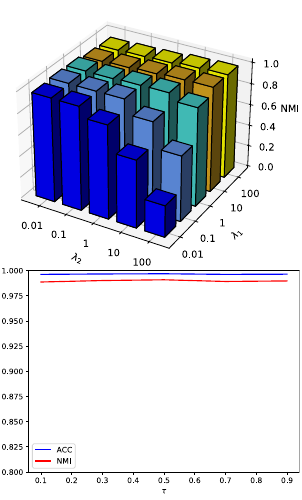}
        \caption{MNIST-USPS}
        \label{fig_param_mnist}
    \end{subfigure}
    \begin{subfigure}[b]{0.49\linewidth}
        \centering
        \includegraphics[width=\linewidth]{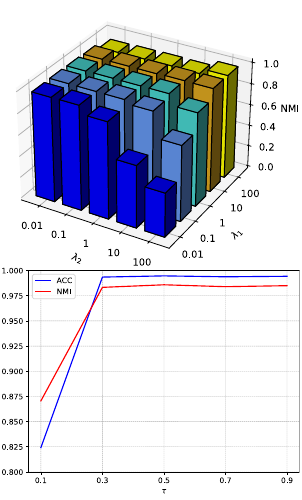}
        \caption{Fashion}
        \label{fig_param_fashion}
    \end{subfigure}
    \caption{Parameter analysis on two datasets.}
    \label{fig_param}
\end{figure}

\subsection{Parameter Sensitivity Analysis}
In this section, we analyze the impact of parameters on BDCL. Fig.~\ref{fig_param} shows the effect of the hyperparameter values in Eq.~\eqref{eq_total_loss} and Eq.~\eqref{eq_contrastive} on the clustering performance. $\lambda_1$ and $\lambda_2$ are used to control the weights of cluster consistency and bi-level decoupling, respectively. The value of $\lambda_1$ has little effect on the final result. However, when the ratio of $\lambda_2/\lambda_1$ is too large, the model tends to learn decoupled feature representations, destroying the original structure of the features, resulting in a sharp drop in clustering performance. In most combinations, the clustering results show good performance and stability. However, when there is an imbalance between the two parameters, the performance will drop significantly. The value of the temperature parameter $\tau$ has little effect on the results. Based on experience, this paper sets $\tau = 0.5$.

\begin{figure}
    \centering
    \begin{subfigure}[b]{0.49\linewidth}
        \centering
        \includegraphics[width=\linewidth]{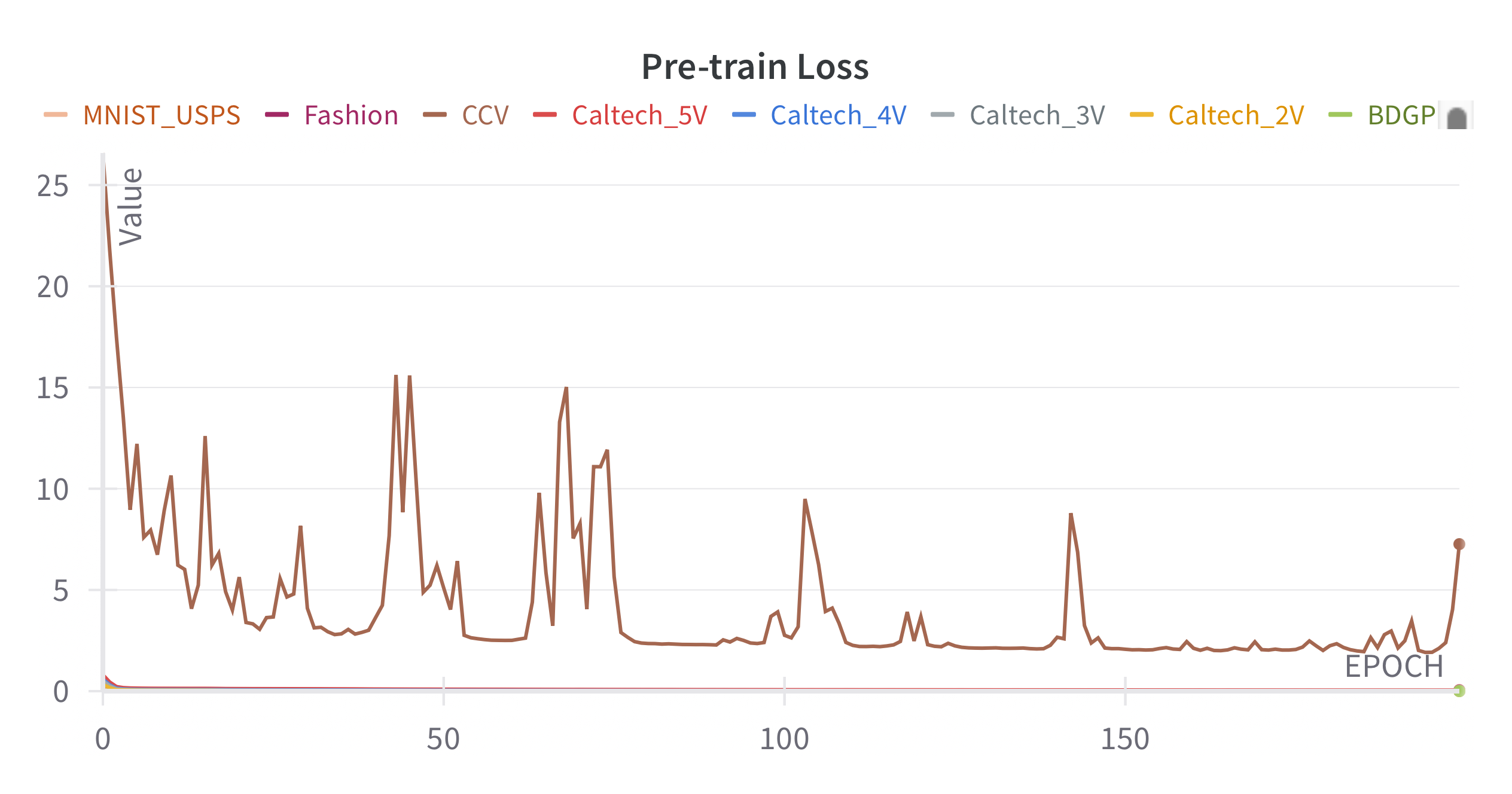}
        \caption{Pre-training phase}
        \label{fig_Pre_tain}
    \end{subfigure}
    \begin{subfigure}[b]{0.49\linewidth}
        \centering
        \includegraphics[width=\linewidth]{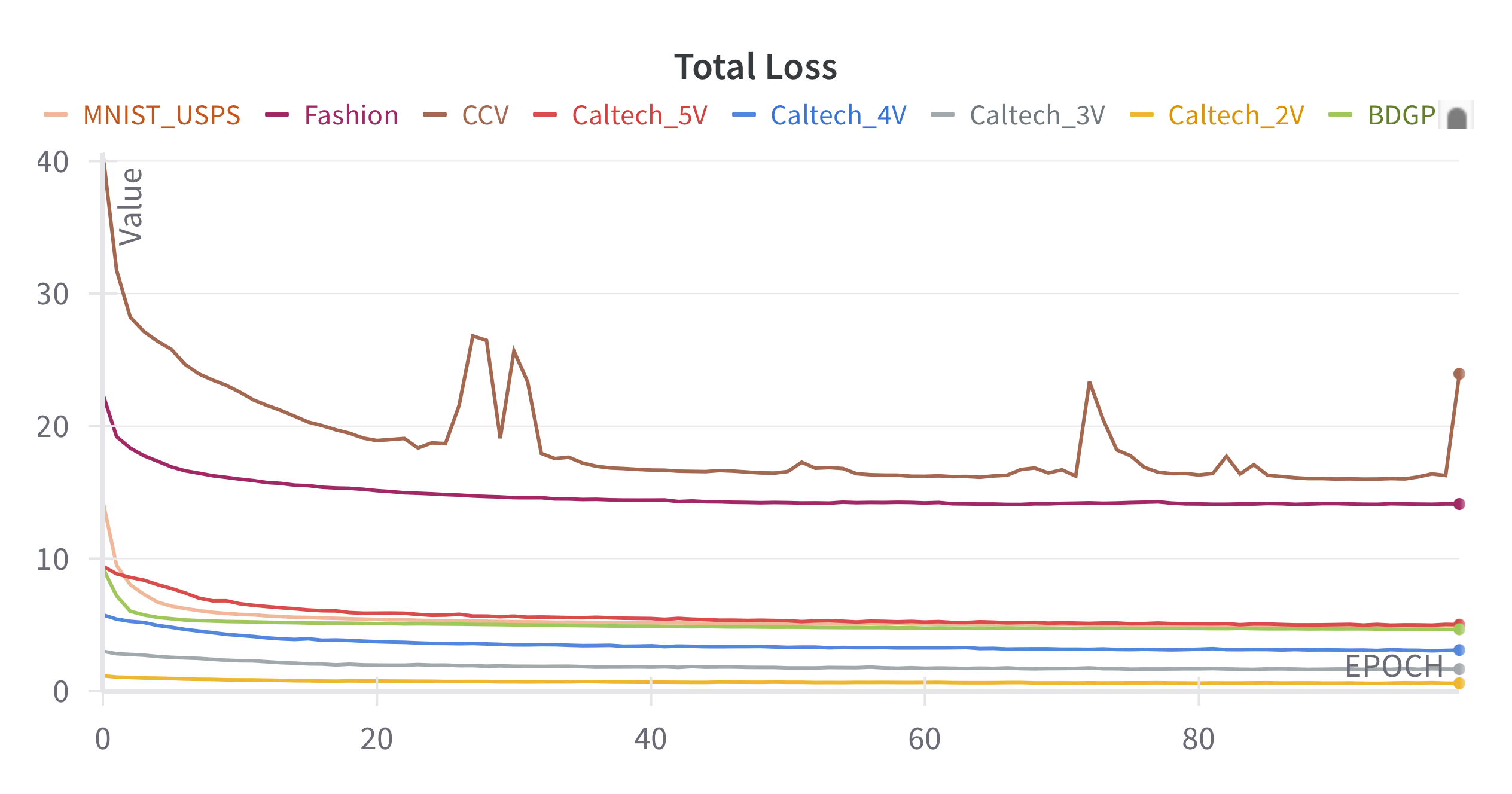}
        \caption{Clustering phase}
        \label{fig_Total_loss}
    \end{subfigure}
    \caption{Loss curves on all datasets.}
    \label{fig_convergence}
\end{figure}
\subsection{Convergence Analysis}
The feature $\mathbf{Z}$ learned by the AE tends to retain the original data $\mathbf{X}$ information to the greatest extent, while contrastive learning tends to learn the discriminative information of sample features. Applying both to the embedded feature $\mathbf{Z}$ will lead to an inevitable conflict, which will result in the loss of feature information. Therefore, we divide the training into a pre-training phase and a clustering phase. The loss function in the pre-training phase is calculated by Eq.~\eqref{eq_reconstruction}, and the loss in the clustering phase is calculated by Eq.~\eqref{eq_total_loss}. As shown in the Fig.~\ref{fig_convergence}, most of the datasets have good convergence. While CCV reconstruction loss is large and fluctuating greatly, the overall trend is convergent. We suppose this is due to the larger dimensionality of the CCV data, causing AE to lose more information when reconstructing the original data.

\begin{figure}
    \centering
    \begin{subfigure}[b]{0.8\linewidth}
        \centering
        \includegraphics[width=\linewidth]{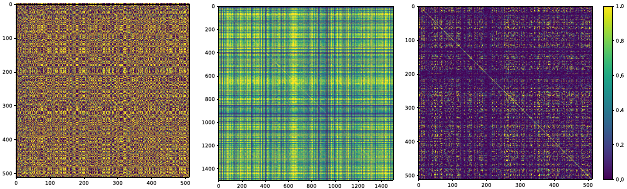}
        \caption{Relationship metrics between embedding features}
        \label{fig_embedding_metrics}
    \end{subfigure}
    \begin{subfigure}[b]{0.8\linewidth}
        \centering
        \includegraphics[width=\linewidth]{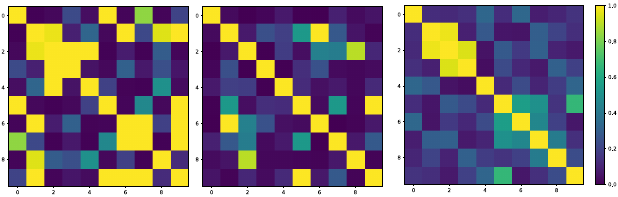}
        \caption{Relationship metrics between cluster features}
        \label{fig_cluster_metrics}
    \end{subfigure}
    \caption{Relationship metrics of different methods on MNIST-USPS. Darker to brighter colors indicate lower to higher degrees of coupling. From left to right are MFL, CVCL and BDCL.}
    \label{fig_metrics}
\end{figure}

\begin{figure}
    \centering
    \begin{subfigure}[b]{0.95\linewidth}
        \centering
        \includegraphics[width=\linewidth]{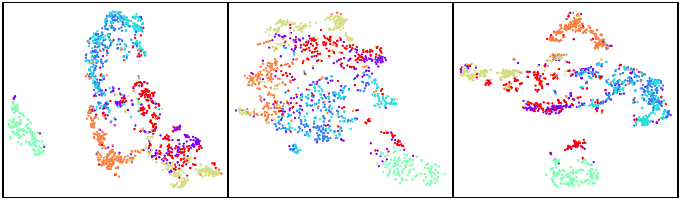}
        \caption{$t$-SNE on Caltech-2V}
        \label{fig_Caltech-2V}
    \end{subfigure}
    \begin{subfigure}[b]{0.95\linewidth}
        \centering
        \includegraphics[width=\linewidth]{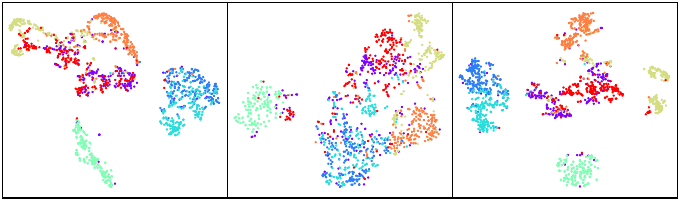}
        \caption{$t$-SNE on Caltech-5V}
        \label{fig_Caltech-5V}
    \end{subfigure}
    \caption{2D $t$-SNE visualization on two datasets. From left to right are MFL, CVCL and BDCL.}
    \label{fig_tsne}
\end{figure}

\subsection{Visualization Analysis}
To further demonstrate the effect of our bi-level decoupling, we visualize the degree of coupling between features and compare it with two popular methods. At the feature-level, as depicted in Fig.~\ref{fig_embedding_metrics}, MFL and CVCL exhibit high coupling, while BDCL effectively decouples features. Meanwhile, at the cluster-level, as illustrated in Fig.~\ref{fig_cluster_metrics}, BDCL demonstrates a lower degree of coupling compared to MFL. CVCL also performs well, an observation from the fact that CVCL promotes cluster features to quadratic values before contrastive learning.

We use $t$-SNE to visualize the trained embedding features in 2D, further analyze the distribution in the embedding features space, and compare them with two popular methods. As depicted in Fig.~\ref{fig_tsne}, the feature distribution of BDCL exhibits greater compactness and higher discrimination between different clusters. Notably, BDCL performs particularly well in cluster separation. This can be attributed to the effective bi-level decoupling of embedding and cluster features, as well as the compression of cluster space, which both enhance the distinguishability of features and the compactness of clusters.

\section{Conclusion}
To further explore effective representations for multi-view data, we propose a novel bi-level decoupling and consistency learning framework for learning representation for DMVC. The bi-level decoupling of feature and cluster aims to reduce their coupling, thereby enhancing feature and cluster discriminability. Additionally, clustering consistency learning is employed to compress the distance between samples and their neighbors, further improving the compactness of the cluster space. Extensive experiments conducted on five benchmark datasets demonstrate that our method achieves SOTA performance.

\clearpage

\bibliographystyle{IEEEbib}
\bibliography{ijcnn2025references}

\begin{thebibliography}{10}

\bibitem{gao2015multi}
Hongchang Gao, Feiping Nie, Xuelong Li, and Heng Huang,
\newblock ``Multi-view subspace clustering,''
\newblock in {\em Proceedings of the IEEE international conference on computer vision}, 2015, pp. 4238--4246.

\bibitem{cao2015diversity}
Xiaochun Cao, Changqing Zhang, Huazhu Fu, Si~Liu, and Hua Zhang,
\newblock ``Diversity-induced multi-view subspace clustering,''
\newblock in {\em Proceedings of the IEEE conference on computer vision and pattern recognition}, 2015, pp. 586--594.

\bibitem{kang2020large}
Zhao Kang, Wangtao Zhou, Zhitong Zhao, Junming Shao, Meng Han, and Zenglin Xu,
\newblock ``Large-scale multi-view subspace clustering in linear time,''
\newblock in {\em Proceedings of the AAAI conference on artificial intelligence}, 2020, vol.~34, pp. 4412--4419.

\bibitem{liu2013multi}
Jialu Liu, Chi Wang, Jing Gao, and Jiawei Han,
\newblock ``Multi-view clustering via joint nonnegative matrix factorization,''
\newblock in {\em Proceedings of the 2013 SIAM international conference on data mining}, 2013, pp. 252--260.

\bibitem{zhao2017multi}
Handong Zhao, Zhengming Ding, and Yun Fu,
\newblock ``Multi-view clustering via deep matrix factorization,''
\newblock in {\em Proceedings of the AAAI conference on artificial intelligence}, 2017, vol.~31.

\bibitem{huang2020auto}
Shudong Huang, Zhao Kang, and Zenglin Xu,
\newblock ``Auto-weighted multi-view clustering via deep matrix decomposition,''
\newblock {\em Pattern Recognition}, vol. 97, pp. 107015, 2020.

\bibitem{tang2020cgd}
Chang Tang, Xinwang Liu, Xinzhong Zhu, En~Zhu, Zhigang Luo, Lizhe Wang, and Wen Gao,
\newblock ``Cgd: Multi-view clustering via cross-view graph diffusion,''
\newblock in {\em Proceedings of the AAAI conference on artificial intelligence}, 2020, vol.~34, pp. 5924--5931.

\bibitem{wang2019gmc}
Hao Wang, Yan Yang, and Bing Liu,
\newblock ``Gmc: Graph-based multi-view clustering,''
\newblock {\em IEEE Transactions on Knowledge and Data Engineering}, vol. 32, no. 6, pp. 1116--1129, 2019.

\bibitem{xie2016unsupervised}
Junyuan Xie, Ross Girshick, and Ali Farhadi,
\newblock ``Unsupervised deep embedding for clustering analysis,''
\newblock in {\em International conference on machine learning}, 2016, pp. 478--487.

\bibitem{guo2017improved}
Xifeng Guo, Long Gao, Xinwang Liu, and Jianping Yin,
\newblock ``Improved deep embedded clustering with local structure preservation.,''
\newblock in {\em Ijcai}, 2017, vol.~17, pp. 1753--1759.

\bibitem{abavisani2018deep}
Mahdi Abavisani and Vishal~M Patel,
\newblock ``Deep multimodal subspace clustering networks,''
\newblock {\em IEEE Journal of Selected Topics in Signal Processing}, vol. 12, no. 6, pp. 1601--1614, 2018.

\bibitem{li2019deep}
Zhaoyang Li, Qianqian Wang, Zhiqiang Tao, Quanxue Gao, Zhaohua Yang, et~al.,
\newblock ``Deep adversarial multi-view clustering network.,''
\newblock in {\em IJCAI}, 2019, vol.~2, p.~4.

\bibitem{xu2021multi}
Jie Xu, Yazhou Ren, Huayi Tang, Xiaorong Pu, Xiaofeng Zhu, Ming Zeng, and Lifang He,
\newblock ``Multi-vae: Learning disentangled view-common and view-peculiar visual representations for multi-view clustering,''
\newblock in {\em Proceedings of the IEEE/CVF international conference on computer vision}, 2021, pp. 9234--9243.

\bibitem{DBLP:conf/icncc/TangTWFW18}
Xiaoliang Tang, Xuan Tang, Wanli Wang, Li~Fang, and Xian Wei,
\newblock ``Deep multi-view sparse subspace clustering,''
\newblock in {\em Proceedings of the {VII} International Conference on Network, Communication and Computing, {ICNCC} 2018, Taipei City, Taiwan, December 14-16, 2018}, 2018, pp. 115--119.

\bibitem{cheng2021multi}
Jiafeng Cheng, Qianqian Wang, Zhiqiang Tao, Deyan Xie, and Quanxue Gao,
\newblock ``Multi-view attribute graph convolution networks for clustering,''
\newblock in {\em Proceedings of the twenty-ninth international conference on international joint conferences on artificial intelligence}, 2021, pp. 2973--2979.

\bibitem{He0WXG20}
Kaiming He, Haoqi Fan, Yuxin Wu, Saining Xie, and Ross~B. Girshick,
\newblock ``Momentum contrast for unsupervised visual representation learning,''
\newblock in {\em 2020 {IEEE/CVF} Conference on Computer Vision and Pattern Recognition, {CVPR} 2020, Seattle, WA, USA, June 13-19, 2020}, 2020, pp. 9726--9735.

\bibitem{ChenK0H20}
Ting Chen, Simon Kornblith, Mohammad Norouzi, and Geoffrey~E. Hinton,
\newblock ``A simple framework for contrastive learning of visual representations,''
\newblock in {\em Proceedings of the 37th International Conference on Machine Learning, {ICML} 2020, 13-18 July 2020, Virtual Event}, 2020, vol. 119, pp. 1597--1607.

\bibitem{xu2022self}
Jie Xu, Yazhou Ren, Huayi Tang, Zhimeng Yang, Lili Pan, Yang Yang, Xiaorong Pu, S~Yu Philip, and Lifang He,
\newblock ``Self-supervised discriminative feature learning for deep multi-view clustering,''
\newblock {\em IEEE Transactions on Knowledge and Data Engineering}, 2022.

\bibitem{yan2023gcfagg}
Weiqing Yan, Yuanyang Zhang, Chenlei Lv, Chang Tang, Guanghui Yue, Liang Liao, and Weisi Lin,
\newblock ``Gcfagg: Global and cross-view feature aggregation for multi-view clustering,''
\newblock in {\em Proceedings of the IEEE/CVF Conference on Computer Vision and Pattern Recognition}, 2023, pp. 19863--19872.

\bibitem{dong2023cross}
Zhibin Dong, Siwei Wang, Jiaqi Jin, Xinwang Liu, and En~Zhu,
\newblock ``Cross-view topology based consistent and complementary information for deep multi-view clustering,''
\newblock in {\em Proceedings of the IEEE/CVF International Conference on Computer Vision}, 2023, pp. 19440--19451.

\bibitem{0001PCZCP024}
Yazhou Ren, Jingyu Pu, Chenhang Cui, Yan Zheng, Xinyue Chen, Xiaorong Pu, and Lifang He,
\newblock ``Dynamic weighted graph fusion for deep multi-view clustering,''
\newblock in {\em Proceedings of the Thirty-Third International Joint Conference on Artificial Intelligence}. 2024, pp. 4842--4850, ijcai.org.

\bibitem{xu2019adversarial}
Cai Xu, Ziyu Guan, Wei Zhao, Hongchang Wu, Yunfei Niu, and Beilei Ling,
\newblock ``Adversarial incomplete multi-view clustering.,''
\newblock in {\em IJCAI}, 2019, vol.~7, pp. 3933--3939.

\bibitem{TaoTN21}
Yaling Tao, Kentaro Takagi, and Kouta Nakata,
\newblock ``Clustering-friendly representation learning via instance discrimination and feature decorrelation,''
\newblock in {\em 9th International Conference on Learning Representations, {ICLR} 2021, Virtual Event, Austria, May 3-7, 2021}. OpenReview.net.

\bibitem{GongZTL22}
Lei Gong, Sihang Zhou, Wenxuan Tu, and Xinwang Liu,
\newblock ``Attributed graph clustering with dual redundancy reduction,''
\newblock in {\em Proceedings of the Thirty-First International Joint Conference on Artificial Intelligence, {IJCAI} 2022, Vienna, Austria, 23-29 July 2022}, Luc~De Raedt, Ed. pp. 3015--3021, ijcai.org.

\bibitem{LiuTZLSYZ22}
Yue Liu, Wenxuan Tu, Sihang Zhou, Xinwang Liu, Linxuan Song, Xihong Yang, and En~Zhu,
\newblock ``Deep graph clustering via dual correlation reduction,''
\newblock in {\em Thirty-Sixth {AAAI} Conference on Artificial Intelligence}. pp. 7603--7611, {AAAI} Press.

\bibitem{xu2022multi}
Jie Xu, Huayi Tang, Yazhou Ren, Liang Peng, Xiaofeng Zhu, and Lifang He,
\newblock ``Multi-level feature learning for contrastive multi-view clustering,''
\newblock in {\em Proceedings of the IEEE/CVF Conference on Computer Vision and Pattern Recognition}, 2022, pp. 16051--16060.

\bibitem{wen2021cdimc}
Jie Wen, Zheng Zhang, Yong Xu, Bob Zhang, Lunke Fei, and Guo-Sen Xie,
\newblock ``Cdimc-net: Cognitive deep incomplete multi-view clustering network,''
\newblock in {\em Proceedings of the Twenty-Ninth International Conference on International Joint Conferences on Artificial Intelligence}, 2021, pp. 3230--3236.

\bibitem{zhou2020end}
Runwu Zhou and Yi-Dong Shen,
\newblock ``End-to-end adversarial-attention network for multi-modal clustering,''
\newblock in {\em Proceedings of the IEEE/CVF conference on computer vision and pattern recognition}, 2020, pp. 14619--14628.

\bibitem{trosten2021reconsidering}
Daniel~J Trosten, Sigurd Lokse, Robert Jenssen, and Michael Kampffmeyer,
\newblock ``Reconsidering representation alignment for multi-view clustering,''
\newblock in {\em Proceedings of the IEEE/CVF conference on computer vision and pattern recognition}, 2021, pp. 1255--1265.

\bibitem{tang2022deep}
Huayi Tang and Yong Liu,
\newblock ``Deep safe multi-view clustering: Reducing the risk of clustering performance degradation caused by view increase,''
\newblock in {\em Proceedings of the IEEE/CVF Conference on Computer Vision and Pattern Recognition}, 2022, pp. 202--211.

\bibitem{Chen_2023_ICCV}
Jie Chen, Hua Mao, Wai~Lok Woo, and Xi~Peng,
\newblock ``Deep multiview clustering by contrasting cluster assignments,''
\newblock in {\em Proceedings of the IEEE/CVF International Conference on Computer Vision (ICCV)}, October 2023, pp. 16752--16761.

\bibitem{ZhangZSCLZ24}
Qian Zhang, Lin Zhang, Ran Song, Runmin Cong, Yonghuai Liu, and Wei Zhang,
\newblock ``Learning common semantics via optimal transport for contrastive multi-view clustering,''
\newblock {\em {IEEE} Trans. Image Process.}, vol. 33, pp. 4501--4515, 2024.

\bibitem{peng2019comic}
Xi~Peng, Zhenyu Huang, Jiancheng Lv, Hongyuan Zhu, and Joey~Tianyi Zhou,
\newblock ``Comic: Multi-view clustering without parameter selection,''
\newblock in {\em International conference on machine learning}. PMLR, 2019, pp. 5092--5101.

\bibitem{cai2012joint}
Xiao Cai, Hua Wang, Heng Huang, and Chris Ding,
\newblock ``Joint stage recognition and anatomical annotation of drosophila gene expression patterns,''
\newblock {\em Bioinformatics}, vol. 28, no. 12, pp. i16--i24, 2012.

\bibitem{jiang2011consumer}
Yu-Gang Jiang, Guangnan Ye, Shih-Fu Chang, Daniel Ellis, and Alexander~C Loui,
\newblock ``Consumer video understanding: A benchmark database and an evaluation of human and machine performance,''
\newblock in {\em Proceedings of the 1st ACM International Conference on Multimedia Retrieval}, 2011, pp. 1--8.

\bibitem{xiao2017fashion}
Han Xiao, Kashif Rasul, and Roland Vollgraf,
\newblock ``Fashion-mnist: a novel image dataset for benchmarking machine learning algorithms,''
\newblock {\em arXiv preprint arXiv:1708.07747}, 2017.

\bibitem{fei2004learning}
Li~Fei-Fei, Rob Fergus, and Pietro Perona,
\newblock ``Learning generative visual models from few training examples: An incremental bayesian approach tested on 101 object categories,''
\newblock in {\em 2004 conference on computer vision and pattern recognition workshop}. IEEE, 2004, pp. 178--178.

\end{thebibliography}

\end{document}